\documentclass{article}

 \usepackage[preprint]{neurips_2026}

\usepackage[utf8]{inputenc} 
\usepackage[T1]{fontenc}    
\usepackage{hyperref}       
\usepackage{url}            
\usepackage{booktabs}       
\usepackage{amsfonts}       
\usepackage{nicefrac}       
\usepackage{microtype}      
\usepackage{xcolor}         
\usepackage{graphicx}
\usepackage{subcaption}
\usepackage{amsmath}
\usepackage{multirow}
\usepackage{siunitx}
\usepackage{float}
\usepackage{makecell}
\usepackage{algorithm}
\title{Sparse Autoencoders as Plug-and-Play Firewalls for Adversarial Attack Detection in VLMs}
\author{Hao Wang$^{1,2}$\thanks{Work done during internship at MTRI.} \quad Yiqun Sun$^{1}$ \quad Pengfei Wei$^{1}$ \quad Lawrence B. Hsieh$^{1}$ \quad Daisuke Kawahara$^{2}$ \\[3pt]
$^{1}$Magellan Technology Research Institute (MTRI) \qquad $^{2}$Waseda University \\[3pt]
\normalsize\texttt{conan1024hao@akane.waseda.jp} \\
\normalsize\texttt{\{duke.sun, pengfei.wei, lawrence.hsieh\}@mtri.co.jp} \\
\normalsize\texttt{dkw@waseda.jp} \\[3pt]
\normalsize Code: \url{https://github.com/conan1024hao/SAEgis}
}

\begin{document}
\maketitle
\begin{abstract}
Vision-language models (VLMs) have advanced rapidly and are increasingly deployed in real-world applications, especially with the rise of agent-based systems.
However, their safety has received relatively limited attention.
Even the latest proprietary and open-weight VLMs remain highly vulnerable to adversarial attacks, leaving downstream applications exposed to significant risks.
In this work, we propose a novel and lightweight adversarial attack detection framework based on sparse autoencoders (SAEs), termed \textbf{SAEgis}.
By inserting an SAE module into a pretrained VLM and training it with standard reconstruction objectives, we find that the learned sparse latent features naturally capture attack-relevant signals.
These features enable reliable classification of whether an input image has been adversarially perturbed, even for previously unseen samples.
Extensive experiments show that SAEgis achieves strong performance across in-domain, cross-domain, and cross-attack settings, with particularly large improvements in cross-domain generalization compared to existing baselines.
In addition, combining signals from multiple layers further improves robustness and stability.
To the best of our knowledge, this is the first work to explore SAE as a plug-and-play mechanism for adversarial attack detection in VLMs.
Our method requires no additional adversarial training, introduces minimal overhead, and provides a practical approach for improving the safety of real-world VLM systems.
\end{abstract}
\begin{figure*}[t]
  \centering
  \includegraphics[width=\linewidth]{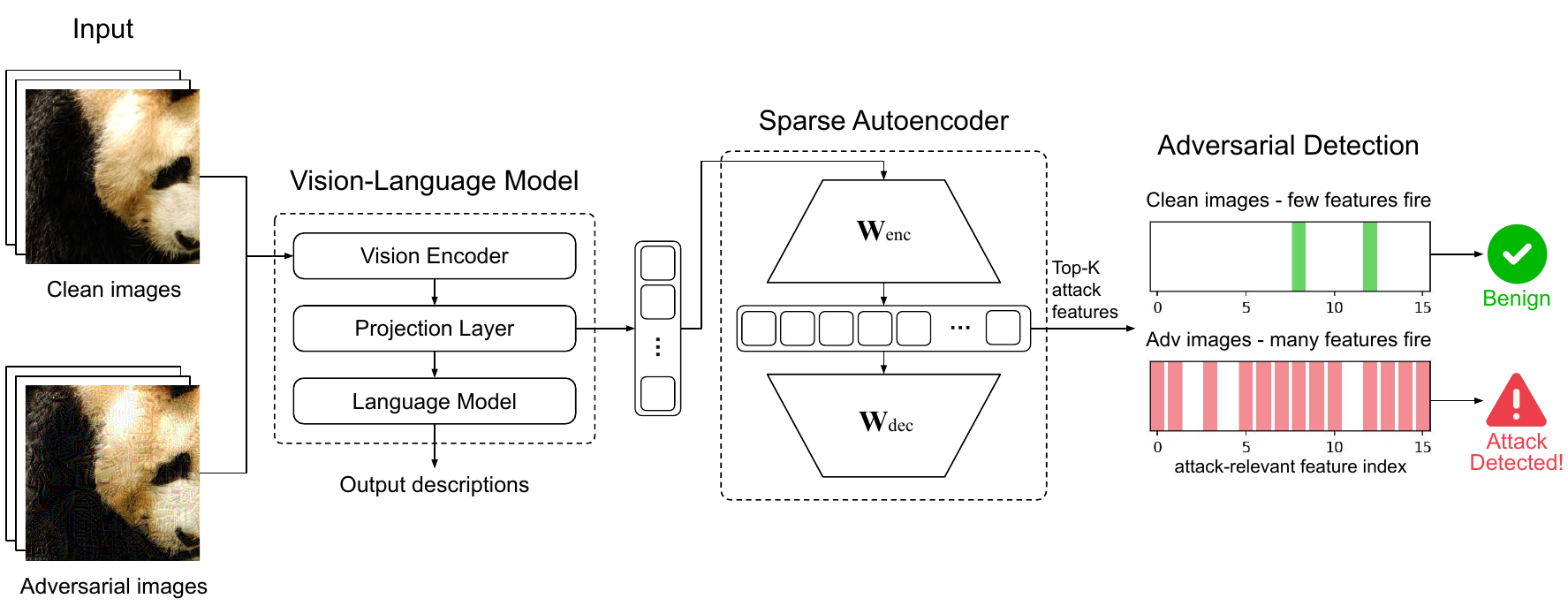}
  \caption{Overview of SAEgis. An SAE is inserted into the VLM and trained with reconstruction. Top-$k$ attack-relevant features are identified from a set of adversarial samples. At inference, inputs activating many such features are flagged as adversarial, while those with few are classified as clean.}
  \label{fig:main-figure}
  \vspace{-5pt}
\end{figure*}
\vspace{-5pt}
\section{Introduction}
\vspace{-5pt}
Vision-language models (VLMs) have advanced rapidly in recent years~\citep{gemmateam2025gemma3technicalreport,nvidia2025nvidianemotronnanov2,clark2026molmo2openweightsdata,vteam2026glm45vglm41vthinkingversatilemultimodal,kimiteam2026kimik25visualagentic,bai2025qwen3vltechnicalreport,qwen3.5}, evolving from early tasks such as visual question answering~\citep{agrawal2016vqavisualquestionanswering}, image captioning~\citep{herdade2020imagecaptioningtransformingobjects}, and visual grounding~\citep{qiao2020referringexpressioncomprehensionsurvey} to more recent capabilities including visual reasoning~\citep{Chen_Zhou_Shen_Hong_Sun_Gutfreund_Gan_2024,thawakar-etal-2025-llamav} and embodied AI~\citep{jiang2025surveyvisionlanguageactionmodelsautonomous,zhang2026vlm4vlarevisitingvisionlanguagemodelsvisionlanguageaction,zhang2026chainofactiontrajectoryautoregressivemodeling}.
As a result, VLMs have transformed from simple image-description chatbots into increasingly indispensable assistants in real-world applications.
Despite these achievements, their safety has not received commensurate attention~\citep{lee2025visionlanguagemodelssafewild,liu2025vlmguardsafeguardingvisionlanguagemodels}.
Unlike pure language models, VLMs take images as input, which introduces additional vulnerabilities and makes them more susceptible to adversarial attacks.
Even state-of-the-art VLMs can be easily misled by adversarially perturbed images, often ignoring the original visual semantics and instead generating responses conditioned on the injected perturbations~\citep{dong2023robustgooglesbardadversarial,NEURIPS2023_a97b58c4,li2025frustratinglysimplehighlyeffective,jia2025adversarialattacksclosedsourcemllms}.
This poses significant security risks for the growing number of real-world systems that deploy VLMs without sufficient safeguards.

Early in the development of VLMs, researchers observed that systems such as ChatGPT and Bard are highly vulnerable to adversarial perturbations on images, leading to the proposal of attack methods such as SSA-CWA~\citep{dong2023robustgooglesbardadversarial} and AttackVLM~\citep{NEURIPS2023_a97b58c4}.
Since then, more efficient attack methods have been introduced~\citep{guo2024efficientgenerationtargetedtransferable,zhang2025anyattacklargescaleselfsupervisedadversarial,li2025frustratinglysimplehighlyeffective,jia2025adversarialattacksclosedsourcemllms}.
A recent study~\citep{zhao2026pushingfrontierblackboxlvlm} reports near-100\% attack success rates on advanced systems such as GPT-5~\citep{singh2025openaigpt5card} and Gemini-2.5-Pro~\citep{comanici2025gemini25pushingfrontier}, suggesting that despite growing awareness of this issue, mainstream VLMs remain largely incapable of defending against such attacks.
While several works have explored detecting adversarial attacks~\citep{fares2024mirrorcheckefficientadversarialdefense,Zhang_2024,huang2024effectiveefficientadversarialdetection,jiang2025hiddendetectdetectingjailbreakattacks,ZHOU2026115498}, they share two common limitations: (1) they do not evaluate against the latest and strongest attack methods, making their reported performance insufficient to establish robustness, and (2) they focus on fixed datasets and attack settings, without considering out-of-domain scenarios that better reflect real-world deployment conditions.

In this work, we propose \textbf{S}parse \textbf{A}uto\textbf{E}ncoders as Ae\textbf{gis} (SAEgis) , a simple yet efficient adversarial attack detection framework based on sparse autoencoders (SAEs)~\citep{Olshausen1996,ng2011sparse}.
Our key insight is that training an SAE within a pretrained VLM using a standard reconstruction objective implicitly captures the patterns of clean visual inputs.
As a result, adversarially perturbed images, which deviate from these patterns, tend to activate distinct sets of latent features that correspond to attack-related signals.
Concretely, we insert an SAE module into the vision encoder or projection layer of the VLM and train it using the reconstruction objective.
Using a small set of adversarial samples, we identify the top-$k$ attack-relevant features.
At inference time, we then analyze their activation patterns: inputs with few activated features are classified as clean, while those exceeding a threshold are flagged as adversarial.
The overall workflow of SAEgis is illustrated in Figure~\ref{fig:main-figure}.
Notably, our framework requires no additional adversarial training, is fully plug-and-play, and introduces minimal computational overhead to the original VLM.

Our experiments demonstrate that SAEgis achieves strong performance in detecting state-of-the-art adversarial attacks, not only under in-domain settings but also in more challenging cross-domain and cross-attack scenarios.
In particular, SAEgis achieves significantly better cross-domain generalization compared to existing baselines.
Furthermore, we find that ensembling SAE signals from multiple layers, including both the vision encoder and the projection layer, leads to additional performance gains.
These results highlight the effectiveness and robustness of SAEgis, suggesting that it provides a practical solution for improving the safety of real-world VLM systems.

\vspace{-5pt}
\section{Related Work}
\vspace{-5pt}
\subsection{Adversarial Attacks on VLMs}
\vspace{-5pt}
With the emergence of early VLM systems, which were often accessible only as black boxes, researchers increasingly shifted their focus toward transfer-based attack methods.
AttackVLM~\citep{NEURIPS2023_a97b58c4} represents one of the first works to study black-box attacks on VLMs, where adversarial images generated using models such as CLIP~\citep{radford2021learningtransferablevisualmodels} and BLIP~\citep{pmlr-v162-li22n} were transferred to attack other models like MiniGPT-4~\citep{zhu2023minigpt4enhancingvisionlanguageunderstanding}.
SSA-CWA~\citep{dong2023robustgooglesbardadversarial} improves transferability by combining Spectrum Simulation Attack~\citep{long2022frequencydomainmodelaugmentation} with Common Weakness Attack~\citep{chen2024rethinkingmodelensembletransferbased}.
AdvDiffVLM~\citep{guo2024efficientgenerationtargetedtransferable} leverages diffusion models~\citep{ho2020denoisingdiffusionprobabilisticmodels} to generate adversarial examples more efficiently.
AnyAttack~\citep{zhang2025anyattacklargescaleselfsupervisedadversarial} trains a noise generator via contrastive learning on the LAION-400M dataset~\citep{schuhmann2021laion400mopendatasetclipfiltered} to produce transferable adversarial perturbations.
M-Attack~\citep{li2025frustratinglysimplehighlyeffective} enhances transferability by applying random cropping and resizing to both the original and target images during optimization, while FOA-Attack~\citep{jia2025adversarialattacksclosedsourcemllms} introduces a feature optimal alignment loss that aligns both local and global features, leading to notable performance improvements.

\vspace{-2pt}
\subsection{Adversarial Detections for VLMs}
\vspace{-2pt}
Several works have explored methods for detecting and defending against adversarial attacks on VLMs.
MirrorCheck~\citep{fares2024mirrorcheckefficientadversarialdefense} proposes to reconstruct images from generated captions using Stable Diffusion~\citep{rombach2022highresolutionimagesynthesislatent} and detect attacks by comparing the embeddings of the reconstructed and original images.
PIP~\citep{Zhang_2024} introduces irrelevant probe questions and leverages attention maps to train an SVM~\citep{Cortes1995} for classifying adversarial inputs.
\citet{huang2024effectiveefficientadversarialdetection} construct a new adversarial dataset and learn steering vectors~\citep{subramani2022extractinglatentsteeringvectors} that capture attack directions, while HiddenDetect~\citep{jiang2025hiddendetectdetectingjailbreakattacks} similarly defines a refusal vector and detects attacks based on cosine similarity with hidden states.
PromptGuard~\citep{ZHOU2026115498} leverages prompt tuning~\citep{lester2021powerscaleparameterefficientprompt} to enable VLMs to reject harmful inputs.
Despite these efforts, existing methods share common limitations: they are often not evaluated against the latest attack methods, and they typically focus on fixed datasets or attack settings, lacking comprehensive evaluation.
In contrast, our study evaluates against recent strong attacks such as M-Attack and FOA-Attack, and demonstrates the effectiveness of SAEgis under more realistic and challenging settings, including cross-domain and cross-attack generalization.
\vspace{-2pt}
\section{Methodology}
\vspace{-2pt}
In this section, we present how SAEgis identifies attack-relevant features and leverages them to detect adversarially perturbed inputs.
As a prerequisite, we assume access to a pretrained VLM together with an SAE module inserted into the model and trained with a standard reconstruction objective.
The SAE can be placed at different locations within the VLM, including the vision encoder, projection layer, or even the language model, and the detailed training process is described in Sec.~\ref{implementation-details}.
Given this setup, the framework consists of two main stages: feature selection and adversarial detection.
We also introduce an ensemble strategy of SAEs across multiple layers.
\vspace{-2pt}
\subsection{Attack-Relevant Feature Selection}
\vspace{-2pt}
To identify attack-relevant features, we first construct a dataset consisting of both clean and adversarial images.
All images are passed through the VLM equipped with the SAE module, and the activations of the SAE’s sparse latent features are recorded.
In this study, we focus on adversarial attacks targeting image description, the canonical open-ended VLM task and a standard evaluation setting in prior adversarial work.
We accordingly use a fixed text prompt, "Describe this image.", and restrict feature scoring to image tokens.
Let $T$ denote the set of image tokens, and let $a_{i,t}$ represent the activation of the $i$-th SAE feature (with $i \in \{1, \dots, D_{\text{sae}}\}$) at token $t \in T$.

For each feature $i$ on input $x$, we define a feature score that jointly captures both the strength and frequency of its activation across image tokens:
\begin{equation}
\mathrm{score}_i(x) \;=\;
\underbrace{\max_{t \in T} a_{i,t}(x)}_{\text{peak strength}}
\;\cdot\;
\underbrace{\log\!\big(1 + |\{t \in T \mid a_{i,t}(x) > 0\}|\big)}_{\text{spatial extent}}.
\label{eq:score}
\end{equation}
The logarithm balances peak strength against spatial extent: both broadly distributed activations (indicative of global perturbations) and strong, spatially concentrated activations (characteristic of localized attacks) carry useful detection signal, whereas a linear count would let the former dominate.
We compute this score for all clean and adversarial inputs, and take the average over each group.
The attack relevance of feature $i$ is then defined as:
\begin{equation}
\mathrm{attack\_score}_i = \mathbb{E}_{x \sim \mathcal{X}_{\text{attack}}}[\mathrm{score}_i(x)] - \mathbb{E}_{x \sim \mathcal{X}_{\text{clean}}}[\mathrm{score}_i(x)],
\end{equation}
where $\mathcal{X}_{\text{clean}}$ and $\mathcal{X}_{\text{attack}}$ denote the sets of clean and adversarial images, respectively.
Rather than training a classifier on top of the SAE, which would require additional optimization and scale poorly with $D_{\text{sae}}$, we adopt this simple difference-of-means. All features are ranked in descending order according to their attack scores, and the top-$K$ features are selected as attack-relevant features for downstream detection.

\subsection{Adversarial Detection}
In practical deployment, the distribution of adversarial inputs is unknown, making it infeasible to calibrate detection thresholds using adversarial data directly.
Instead, we estimate the threshold solely based on clean data.
To this end, we construct a held-out clean development set and, for each image, compute the number of activated attack-relevant features.
Specifically, given the selected top-$K$ attack-relevant features, we define the activation count for an input $x$ as:
\begin{equation}
\label{eq-3}
N(x)
= \frac{1}{|T|}
\sum_{t \in T}
\sum_{i=1}^{K}
\mathbf{1}\!\left(a_{i,t}(x) > 0\right).
\end{equation}
Intuitively, $N(x)$ measures how many attack-relevant features are triggered by the input image.

We determine the detection threshold based on the empirical distribution of $N(x)$ over the clean development set.
Given a target false positive rate $\alpha$ (e.g., $\alpha = 0.02$), we set the threshold $\tau$ as the $(1 - \alpha)$-quantile:
\begin{equation}
\tau = \mathrm{Quantile}_{1-\alpha} \left( \{ N(x) \mid x \in \mathcal{X}_{\text{clean}}^{\text{dev}} \} \right).
\end{equation}
At inference time, an input is classified as adversarial if $N(x) > \tau$, and as clean otherwise.
This procedure ensures that at most an $\alpha$ fraction of clean samples are falsely flagged as adversarial, providing a reliable way to control the false positive rate in realistic settings.

\subsection{Multi-Layer SAE Ensembling}
\label{sec:ensemble}
Prior work has shown that SAEs trained at different layers of language models capture features with distinct semantic properties~\citep{shi2025routesparseautoencoderinterpret}.
In a VLM this stratification is especially pronounced: early vision layers encode low-level patterns such as textures and edges, while deeper layers encode increasingly global, semantic content.
Adversarial perturbations may surface at any of these levels, with pixel-space noise primarily disrupting early features and semantic or patch-based attacks leaving their cleanest signature deeper in the network, so single-layer detection risks blind spots whose location is itself attack-dependent.

To exploit this complementarity, we extend SAEgis with a simple multi-layer ensemble. Given SAE modules inserted at a set of layers $\mathcal{L}$, we compute the per-layer statistic $N_\ell(x)$ from Eq.~\ref{eq-3} using each layer's own attack-relevant feature set $\mathcal{S}_K^{(\ell)}$, and aggregate them by uniform averaging:
\begin{equation}
\bar{N}(x) = \frac{1}{|\mathcal{L}|} \sum_{\ell \in \mathcal{L}} N_\ell(x).
\end{equation}
The aggregated score $\bar{N}(x)$ is then thresholded exactly as in the single-layer case, with $\bar{\tau}$ chosen as the $(1-\alpha)$-quantile on $\mathcal{X}_{\text{clean}}^{\text{dev}}$, so the clean-only calibration property is preserved end-to-end.
Despite its simplicity, this ensemble improves detection performance and yields more stable behavior across in-domain, cross-domain, and cross-attack settings.
\section{Experiments}
\subsection{Experimental Setup}
\subsubsection{Task Definition and Evaluation}
In this work, we formulate adversarial detection as an image-only binary classification task, where no textual input is provided.
The test set consists of an equal number of clean and adversarial images, and the goal is to determine whether a given input image has been adversarially perturbed.
We set a target level $\alpha = 0.02$ and determine the detection threshold on the clean development set.
We then evaluate the model on the test set by reporting precision, recall, and F1-score under this threshold, providing a standardized comparison across methods at a controlled false positive rate.

We conduct experiments under three evaluation settings: \textit{in-domain}, \textit{cross-domain}, and \textit{cross-attack}.
In the \textit{in-domain} setting, both feature extraction and evaluation are performed on the same dataset.
In the \textit{cross-domain} setting, features are extracted from one dataset while evaluation is conducted on a different dataset, assessing generalization across data distributions.
In the \textit{cross-attack} setting, attack-relevant features are identified using adversarial examples generated by one attack method, while evaluation is performed on adversarial samples produced by a different attack method, measuring robustness to unseen attacks.

\subsubsection{Datasets}
We conduct experiments on three datasets: NIPS17~\citep{nips-2017-non-targeted-adversarial-attack}, LLaVA-Instruct-150K~\citep{liu2023visualinstructiontuning} (LLaVA), and Medical Multimodal Evaluation Data~\citep{chen2024huatuogptvisioninjectingmedicalvisual} (Medical).
The first two consist of natural images, while the third contains medical images for out-of-domain evaluation.
For each dataset, we construct clean splits of 800, 100, and 100 images for training (i.e., feature extraction), development (i.e., threshold calibration), and testing, respectively, and separately generate adversarial examples using 100 images each for training and testing.

\subsubsection{Attack Methods}
We consider three representative adversarial attack methods: SSA-CWA~\citep{dong2023robustgooglesbardadversarial}, M-Attack~\citep{li2025frustratinglysimplehighlyeffective}, and FOA-Attack~\citep{jia2025adversarialattacksclosedsourcemllms}.
SSA-CWA is an earlier, widely used baseline, while M-Attack and FOA-Attack are more recent and stronger, making them suitable for evaluating robustness under advanced threat scenarios.
In the cross-attack setting, we construct evaluation pairs from weaker to stronger attacks.
Specifically, we consider two configurations: SSA-CWA $\rightarrow$ M-Attack and SSA-CWA $\rightarrow$ FOA-Attack, where the source attack is used for feature selection and the target attack is used for evaluation.

\subsubsection{Baseline Approaches}
In addition to SAEgis, we compare against several baselines.
Inspired by~\citet{huang2024effectiveefficientadversarialdetection} and~\citet{jiang2025hiddendetectdetectingjailbreakattacks}, we introduce a simple yet strong \textit{dense baseline}, which operates directly on hidden states rather than sparse latent features.
Specifically, we extract hidden states from a chosen model layer and compute average embeddings for clean and adversarial images.
At inference time, a test image is classified based on its cosine similarity to these two reference representations.
Analogous to SAEgis, we also construct a multi-layer ensemble by aggregating similarity scores across multiple layers.
We also include PIP~\citep{Zhang_2024} as a representative prior method, which trains an SVM classifier using attention maps obtained from irrelevant probe questions to distinguish adversarial inputs.
Since PIP utilizes signals from all language model layers by default, it can also be viewed as an ensemble-based approach.
We additionally evaluated the SAE's reconstruction error (MSE) as a direct anomaly score, but found negligible differences between clean and adversarial inputs; we therefore omit it from our baselines.

\subsection{Implementation Details}
\label{implementation-details}
In this subsection, we describe the implementation details of SAEgis, including SAE pretraining, feature extraction, and threshold calibration.
Qwen2.5-VL-3B-Instruct~\citep{bai2025qwen25vltechnicalreport} is adopted as the backbone VLM.
More recent models such as the Qwen3-VL series~\citep{bai2025qwen3vltechnicalreport} are not used, as their DeepStack architecture~\citep{NEURIPS2024_29cd7f83} injects visual signals into multiple layers of the language model, disrupting the direct propagation of visual information.
To enable clearer analysis of how different layers contribute to adversarial signal detection, we instead choose Qwen2.5-VL, which follows a more straightforward architecture.
We independently train SAE modules at nine different locations within the model, including the vision encoder, projection layer, and language model.
We use the FineVision dataset~\citep{wiedmann2025finevisionopendataneed}, training on 500k samples with a batch size of 16 and a learning rate of 5e-5.
The SAE latent dimensionality is set to 32,768, with a top-$K$ sparsity of 64.
All pretrained SAE weights will be released upon publication.

\begin{figure}[t]
    \centering
    \begin{subfigure}[t]{0.49\textwidth}
        \centering
        \includegraphics[width=\linewidth]{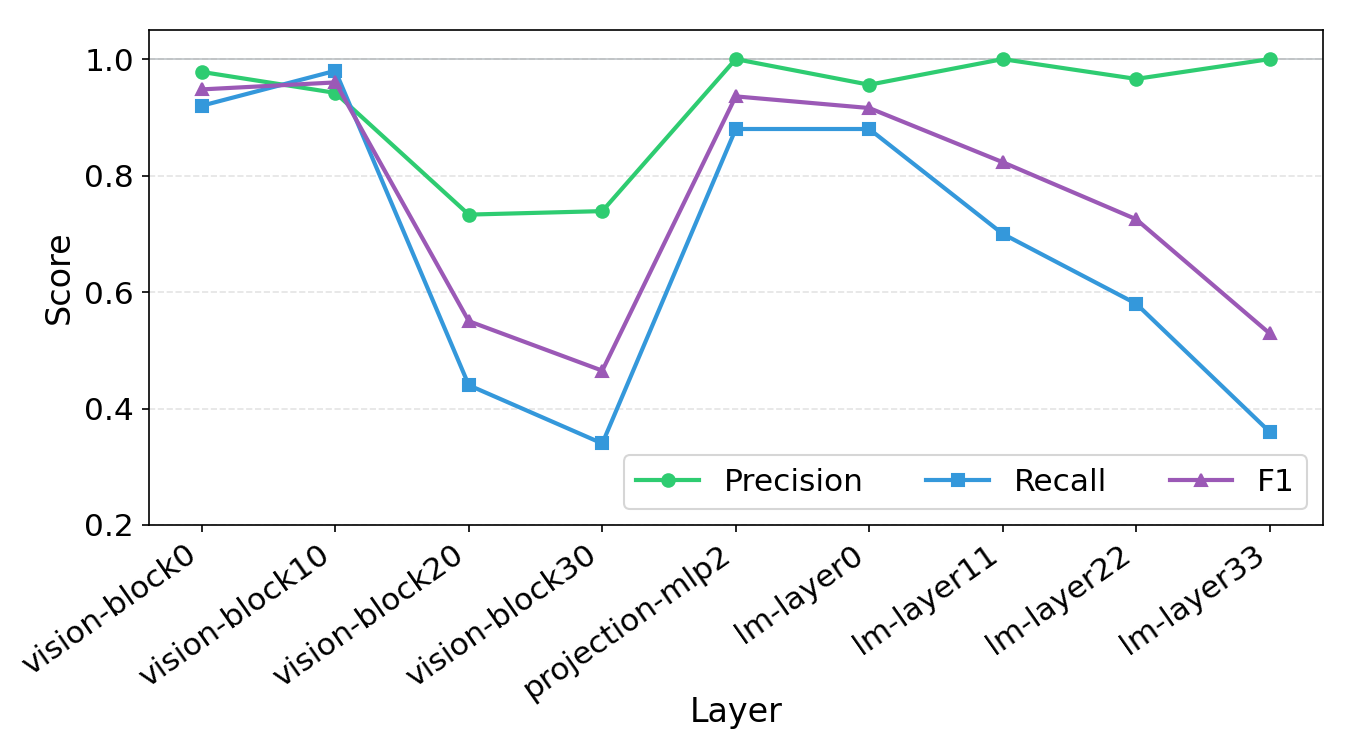}
        \caption{Performance of SAE modules at different locations. The number of selected features is set to $K = 256$.}
        \label{fig:optimal-layer}
    \end{subfigure}
    \hfill
    \begin{subfigure}[t]{0.49\textwidth}
        \centering
        \includegraphics[width=\linewidth]{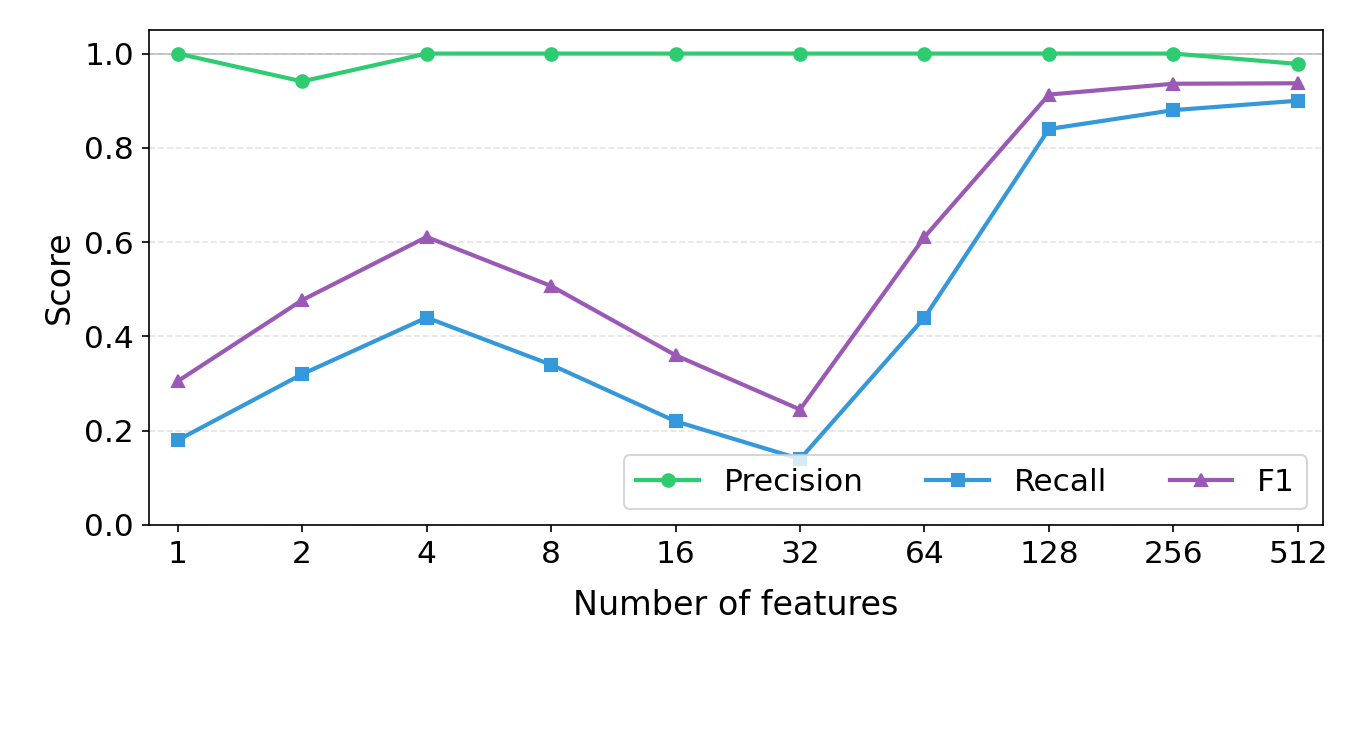}
        \caption{Performance across different values of $K$ for feature selection. \textit{projection-mlp2} is used for SAE insertion.}
        \label{fig:optimal-feature-num}
    \end{subfigure}
    \caption{Preliminary experimental results. Both experiments employ FOA-Attack as the adversarial method and are evaluated under the NIPS17$\rightarrow$Medical transfer setting.}
\end{figure}

In practical deployment of SAEgis, two key design questions arise: (1) which layer is most effective for inserting the SAE module, and (2) what is the optimal number of attack-relevant features (top-$K$)?
To investigate these factors, we conduct a series of preliminary experiments. As shown in Figure~\ref{fig:optimal-layer}, SAEs placed at \textit{vision-block0}, \textit{vision-block10}, and \textit{projection-mlp2} achieve the best performance among the nine candidate locations.
The former two correspond to early vision layers that primarily capture high-frequency patterns such as textures and edges, while the latter serves as a critical interface that projects visual representations into the language model, potentially encoding more global and semantically rich information.
Furthermore, Figure~\ref{fig:optimal-feature-num} suggests that using at least 128 features is necessary to obtain stable recall, indicating that adversarial signals are typically manifested through the joint activation of multiple features rather than isolated ones.
Based on these findings, unless otherwise specified, SAEgis uses the \textit{projection-mlp2} layer for single-layer evaluation in the later experiments, while the ensemble variant aggregates signals from \textit{vision-block0}, \textit{vision-block10}, and \textit{projection-mlp2}.
The same layer configuration is adopted for the dense baseline.
For the number of features, we fix $K=256$ in all subsequent experiments.

\subsection{Main Results}
\label{main-results}
Tables~\ref{tab:in-domain}, \ref{tab:cross-domain}, and \ref{tab:cross-attack} report the performance of all methods under the in-domain, cross-domain, and cross-attack settings, with averaged results summarized in Table~\ref{tab:overall}.
The in-domain and cross-domain averages are taken directly from their respective tables, while the cross-attack results are computed by averaging over each target attack (M-Attack and FOA-Attack) in Table~\ref{tab:cross-attack} and subtracting the corresponding no-transfer scores from Table~\ref{tab:in-domain}.

In the in-domain setting (Table~\ref{tab:in-domain}), all methods achieve strong performance, with Dense (Ensemble) and SAEgis (Ensemble) performing the best overall.
Ensembling signals from multiple layers substantially improves recall across methods, highlighting the benefit of aggregating complementary representations.
SAEgis performs slightly worse than the dense baseline on the Medical dataset, likely because sparse latent features are less expressive than dense hidden states, limiting their advantage in overfitting-friendly scenarios.

In the cross-domain setting (Table~\ref{tab:cross-domain}), we observe that baseline methods face significant challenges in generalization.
The dense baseline, including its ensemble variant, suffers a substantial drop in precision, decreasing from near-perfect performance in the in-domain setting to around 70\% on average.
In contrast, PIP experiences a notable degradation in recall.
SAEgis, however, remains remarkably stable: after ensembling, both precision and recall stay above 90\% in most cases, significantly outperforming other methods.
We further find that both directions of domain shift, namely transferring from common datasets to specialized domains (e.g., LLaVA $\rightarrow$ Medical) and from specialized domains to general ones (e.g., Medical $\rightarrow$ NIPS17/LLaVA), pose serious challenges to model robustness, highlighting the importance of cross-domain evaluation for adversarial detection.

In the cross-attack setting (Table~\ref{tab:cross-attack}), we observe that both the dense baseline and SAEgis relying on the \textit{projection-mlp2} layer suffer significant performance degradation, with the dense baseline even exhibiting near-zero recall on NIPS17.
To address this issue, we additionally evaluate SAEgis using the \textit{vision-block0} layer and find that it maintains strong performance.
We hypothesize that different attack methods induce distinct global feature shifts, making signals from the projection layer less reliable for detection, while low-level perturbations such as high-frequency textures and edges remain more consistent across attacks, allowing early vision layers to better capture these patterns.
We provide a more detailed analysis in Sec.~\ref{analysis:feature}.
Nevertheless, after ensembling,  both the dense baseline and SAEgis recover to performance levels comparable to the no-transfer setting (i.e., in-domain).
While PIP also shows reasonable performance, it is less stable than these two methods.
Interestingly, although the \textit{projection-mlp2} layer performs poorly when used alone, incorporating it into the ensemble still leads to further improvements, a phenomenon we analyze in Sec.~\ref{analysis:ablation}.

Overall, our results demonstrate that SAEgis achieves consistently strong performance across all three evaluation settings.
These findings suggest that features from SAEs capture attack-relevant signals that generalize beyond the specific domain and attack strategy seen during training, making SAEgis a reliable foundation for adversarial detection in real-world VLM deployments under distribution shift.
\begin{table}[t]
\centering
\small
\caption{In-domain results. Highest precision and recall are \underline{underlined}; highest F1 scores are \textbf{bolded}.}
\resizebox{0.77\linewidth}{!}{%
\begin{tabular}{llrrrrrrrrr}
\toprule
\multicolumn{2}{c}{} & \multicolumn{3}{c}{SSA-CWA} & \multicolumn{3}{c}{M-Attack} & \multicolumn{3}{c}{FOA-Attack} \\
\cmidrule(lr){1-2} \cmidrule(lr){3-5} \cmidrule(lr){6-8} \cmidrule(lr){9-11}
Data & Method & \multicolumn{1}{c}{P} & \multicolumn{1}{c}{R} & \multicolumn{1}{c}{F1} & \multicolumn{1}{c}{P} & \multicolumn{1}{c}{R} & \multicolumn{1}{c}{F1} & \multicolumn{1}{c}{P} & \multicolumn{1}{c}{R} & \multicolumn{1}{c}{F1} \\
\midrule
\multirow{5}{*}{NIPS17} & Dense & \underline{100.0} & 89 & 94.1 & \underline{100.0} & 87 & 93.0 & \underline{100.0} & 85 & 91.8 \\
 & Dense (Ensemble) & 99.0 & \underline{100} & 99.5 & 98.0 & \underline{100} & 99.0 & 98.0 & \underline{100} & 99.0 \\
 & PIP & 97.9 & 95 & 96.4 & 97.7 & 87 & 92.0 & 97.7 & 85 & 90.9 \\
 & SAEgis & 97.0 & 98 & 97.5 & 98.9 & 95 & 96.9 & 98.9 & 95 & 96.9 \\
 & SAEgis (Ensemble) & \underline{100.0} & \underline{100} & \textbf{100.0} & 99.0 & \underline{100} & \textbf{99.5} & 99.0 & \underline{100} & \textbf{99.5} \\
\midrule
\multirow{5}{*}{LLaVA} & Dense & 97.0 & 99 & 97.9 & 96.7 & 88 & 92.1 & 96.5 & 83 & 89.2 \\
 & Dense (Ensemble) & \underline{99.0} & \underline{100} & \textbf{99.5} & 93.4 & \underline{100} & 96.6 & 93.4 & \underline{100} & 96.6 \\
 & PIP & 98.0 & \underline{100} & 99.0 & 97.9 & 96 & 96.9 & 97.9 & 94 & 95.9 \\
 & SAEgis & 98.0 & 99 & 98.5 & \underline{98.8} & 86 & 91.9 & 96.5 & 85 & 90.4 \\
 & SAEgis (Ensemble) & 98.0 & \underline{100} & 99.0 & 98.0 & 99 & \textbf{98.5} & \underline{98.0} & 99 & \textbf{98.5} \\
\midrule
\multirow{5}{*}{Medical} & Dense & 98.9 & 90 & 94.2 & 97.7 & 86 & 91.4 & 97.8 & 89 & 93.1 \\
 & Dense (Ensemble) & \underline{100.0} & \underline{97} & \textbf{98.4} & \underline{98.9} & \underline{95} & \textbf{96.9} & \underline{98.9} & \underline{95} & \textbf{96.9} \\
 & PIP & 97.8 & 93 & 95.3 & 97.8 & 90 & 93.7 & 97.8 & 92 & 94.8 \\
 & SAEgis & 98.8 & 88 & 93.1 & 98.7 & 79 & 87.7 & 97.6 & 84 & 90.3 \\
 & SAEgis (Ensemble) & 97.8 & 92 & 94.8 & 94.7 & 91 & 92.8 & 94.8 & 92 & 93.4 \\
\bottomrule
\end{tabular}}
\label{tab:in-domain}
\end{table}
\begin{table}[t]
\centering
\vspace{-10pt}
\caption{Cross-domain results.}
\label{tab:cross-domain}
\small
\resizebox{0.83\linewidth}{!}{%
\begin{tabular}{llrrrrrrrrr}
\toprule
\multicolumn{2}{c}{} & \multicolumn{3}{c}{SSA-CWA} & \multicolumn{3}{c}{M-Attack} & \multicolumn{3}{c}{FOA-Attack} \\
\cmidrule(lr){1-2} \cmidrule(lr){3-5} \cmidrule(lr){6-8} \cmidrule(lr){9-11}
Data & Method & \multicolumn{1}{c}{P} & \multicolumn{1}{c}{R} & \multicolumn{1}{c}{F1} & \multicolumn{1}{c}{P} & \multicolumn{1}{c}{R} & \multicolumn{1}{c}{F1} & \multicolumn{1}{c}{P} & \multicolumn{1}{c}{R} & \multicolumn{1}{c}{F1} \\
\midrule
\multirow{5}{*}{NIPS17$\rightarrow$Medical} & Dense & 95.8 & 93 & 94.3 & \underline{100.0} & 68 & 80.9 & \underline{100.0} & 69 & 81.6 \\
 & Dense (Ensemble) & 92.5 & \underline{100} & \textbf{96.1} & 92.5 & \underline{99} & \textbf{95.6} & 96.1 & \underline{100} & \textbf{98.0} \\
 & PIP & 97.8 & 90 & 93.7 & 97.7 & 88 & 92.6 & 97.6 & 84 & 90.3 \\
 & SAEgis & 84.9 & 90 & 87.3 & 97.5 & 80 & 87.9 & 98.8 & 82 & 89.6 \\
 & SAEgis (Ensemble) & \underline{98.9} & 93 & 95.8 & 98.9 & 90 & 94.2 & 98.9 & 92 & 95.3 \\
\midrule
\multirow{5}{*}{LLaVA$\rightarrow$Medical} & Dense & 50.5 & \underline{100} & 67.1 & 77.1 & 88 & 82.2 & 76.0 & 92 & 83.2 \\
 & Dense (Ensemble) & 59.8 & \underline{100} & 74.9 & 67.5 & \underline{100} & 80.6 & 69.9 & \underline{100} & 82.3 \\
 & PIP & \underline{97.8} & 89 & \textbf{93.1} & 97.7 & 86 & 91.4 & 97.6 & 83 & 89.7 \\
 & SAEgis & 69.1 & 92 & 78.9 & \underline{98.5} & 66 & 79.0 & \underline{98.6} & 72 & 83.2 \\
 & SAEgis (Ensemble) & 88.1 & 97 & 92.3 & 96.8 & 91 & \textbf{93.8} & 97.8 & 93 & \textbf{95.3} \\
\midrule
\multirow{5}{*}{Medical$\rightarrow$NIPS17} & Dense & 61.3 & \underline{100} & 76.0 & 50.7 & \underline{100} & 67.3 & 50.7 & \underline{100} & 67.3 \\
 & Dense (Ensemble) & 83.3 & \underline{100} & 90.9 & 56.5 & \underline{100} & 72.2 & 56.1 & \underline{100} & 71.9 \\
 & PIP & 97.8 & 91 & 94.3 & 96.6 & 57 & 71.7 & \underline{97.1} & 67 & 79.2 \\
 & SAEgis & \underline{98.8} & 87 & 92.5 & \underline{97.9} & 94 & \textbf{95.9} & 95.1 & 98 & \textbf{96.5} \\
 & SAEgis (Ensemble) & 97.0 & \underline{100} & \textbf{98.5} & 84.0 & \underline{100} & 91.3 & 81.9 & \underline{100} & 90.0 \\
\midrule
\multirow{5}{*}{Medical$\rightarrow$LLaVA} & Dense & 51.5 & \underline{100} & 67.9 & 50.0 & \underline{100} & 66.6 & 50.0 & \underline{100} & 66.6 \\
 & Dense (Ensemble) & 80.0 & \underline{100} & 88.8 & 51.2 & \underline{100} & 67.8 & 51.0 & \underline{100} & 67.5 \\
 & PIP & 96.0 & 48 & 64.0 & \underline{94.4} & 34 & 50.0 & \underline{93.3} & 28 & 43.0 \\
 & SAEgis & \underline{100.0} & 78 & 87.6 & 91.8 & 90 & 90.9 & 88.7 & 95 & 91.7 \\
 & SAEgis (Ensemble) & 98.0 & \underline{100} & \textbf{99.0} & 90.8 & 99 & \textbf{94.7} & 86.0 & 99 & \textbf{92.0} \\
\bottomrule
\end{tabular}}
\end{table}

\vspace{-10pt}
\section{Analysis}
\vspace{-5pt}
\subsection{Interpreting Attack-Relevant Features}
\label{analysis:feature}
\begin{figure}[t]
    \centering
    \begin{subfigure}[t]{0.3\textwidth}
        \centering
        \includegraphics[width=\linewidth]{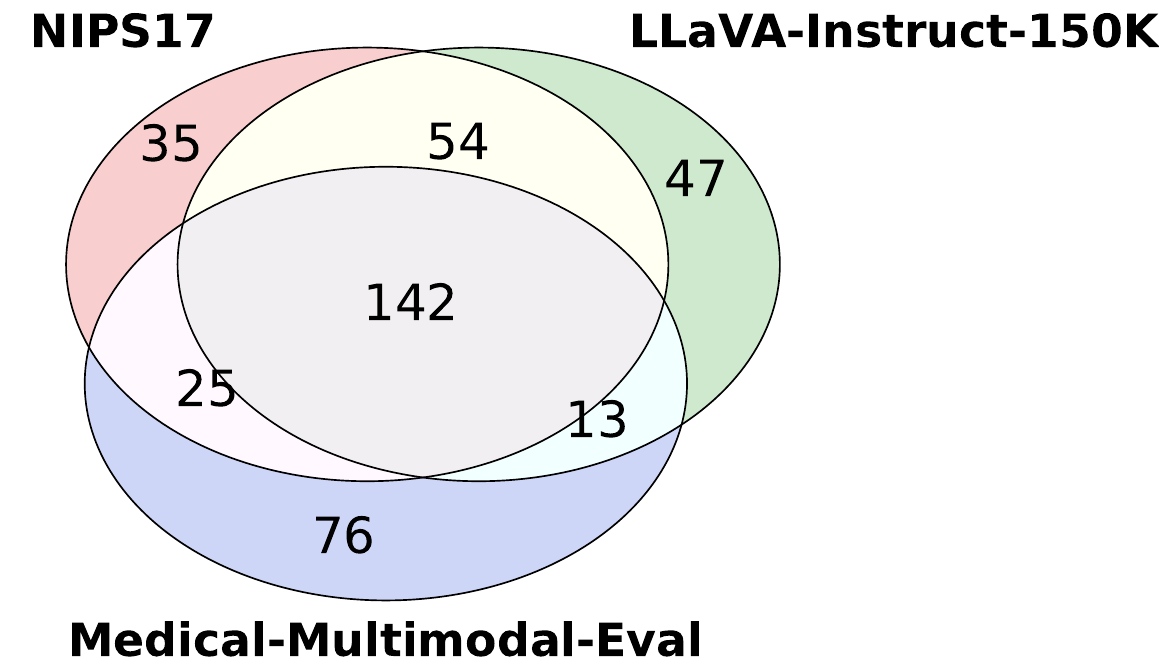}
        \caption{SSA-CWA}
    \end{subfigure}
    \hfill
    \begin{subfigure}[t]{0.3\textwidth}
        \centering
        \includegraphics[width=\linewidth]{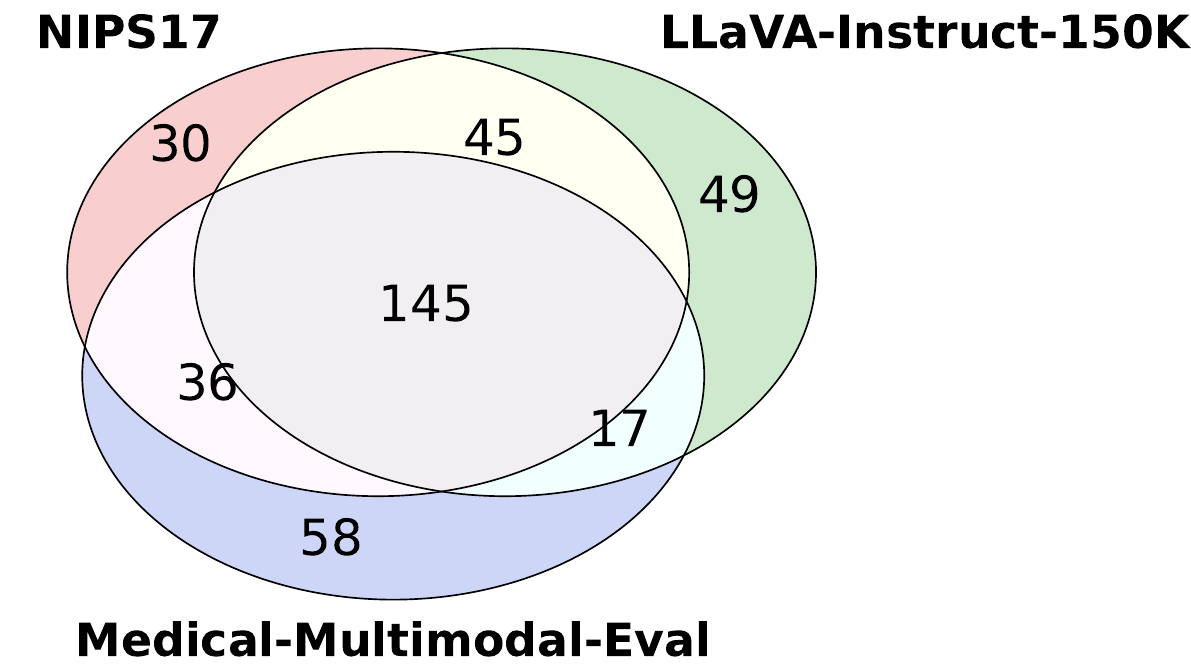}
        \caption{M-Attack}
    \end{subfigure}
    \hfill
    \begin{subfigure}[t]{0.3\textwidth}
        \centering
        \includegraphics[width=\linewidth]{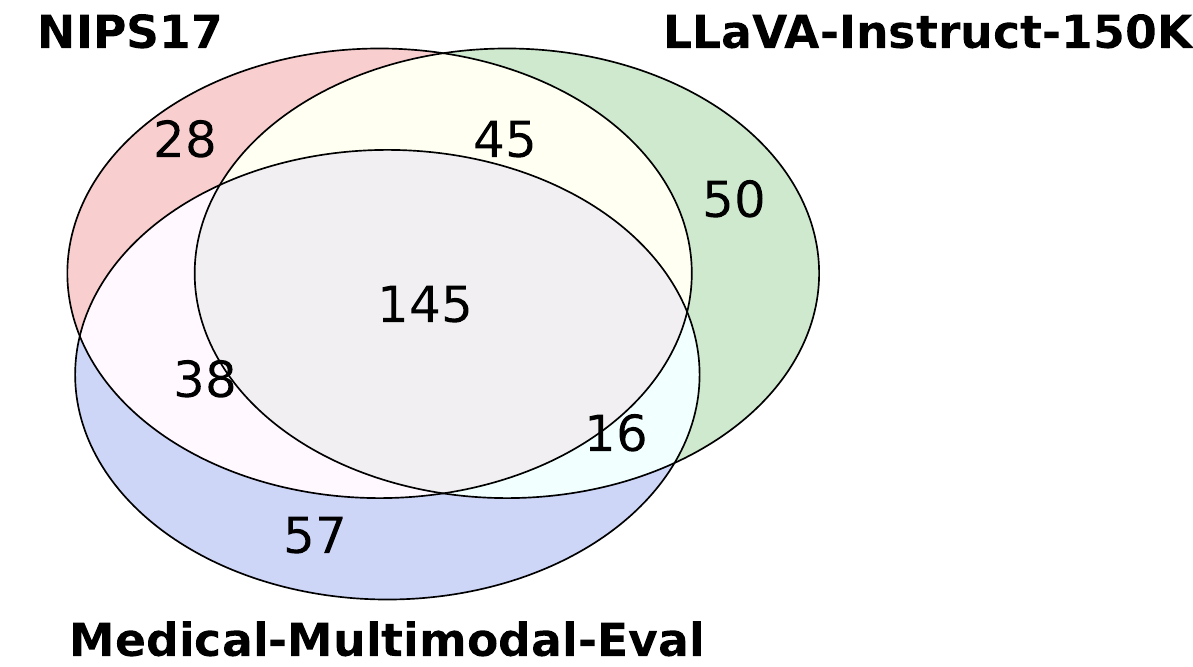}
        \caption{FOA-Attack}
    \end{subfigure}
    \caption{Shared feature overlap across datasets under different attacks, illustrated by Venn diagrams of the top-256 features extracted from three datasets at the \textit{projection-mlp2} layer.}
    \label{fig:feature-venn-cross-domain}
    \vspace{-10pt}
\end{figure}

\paragraph{How are features shared across datasets and attack methods?}
SAEgis achieves strong performance in both cross-domain and cross-attack settings, which raises an important question: are these attack-relevant features consistently activated across different datasets and attack methods?
To investigate this, we visualize the overlap of activated features across three datasets in Figure~\ref{fig:feature-venn-cross-domain}.
We observe that, regardless of the attack method, a substantial portion of features are shared across all datasets.
Notably, even the Medical dataset, which differs significantly in domain from the other two, still shares a large number of features.
This finding suggests that SAEgis captures genuinely transferable attack-relevant representations, providing empirical evidence that its strong cross-domain performance reflects inherent generalizability rather than incidental overlap.

\begin{figure}[t]
    \centering
    \begin{subfigure}[t]{0.3\textwidth}
        \centering
        \includegraphics[width=\linewidth]{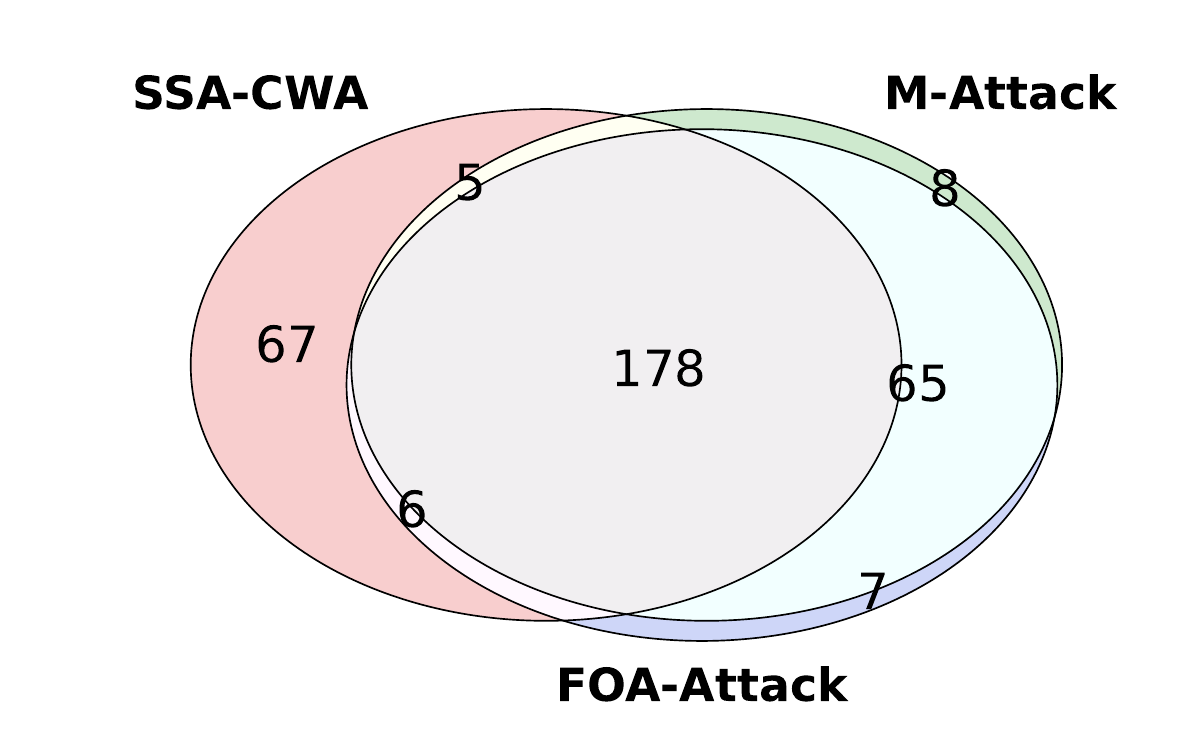}
        \vspace{-15pt}
        \caption{\textit{vision-block0}}
    \end{subfigure}
    \hfill
    \begin{subfigure}[t]{0.3\textwidth}
        \centering
        \includegraphics[width=\linewidth]{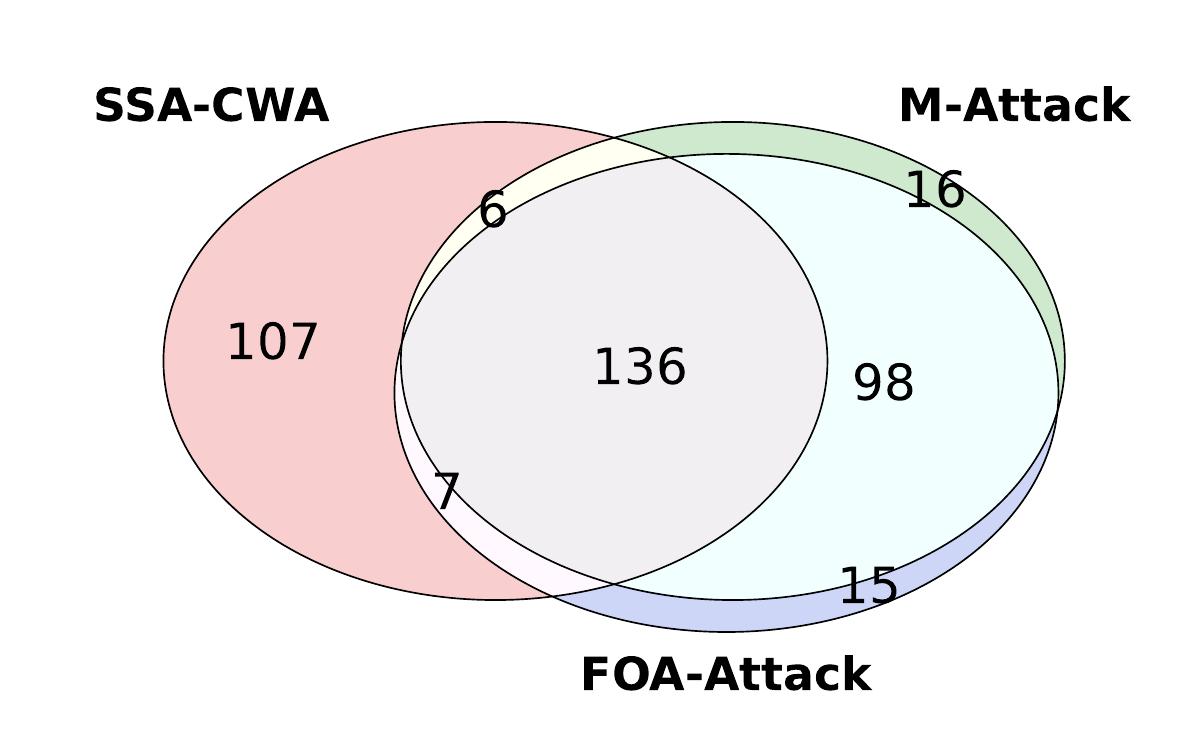}
        \vspace{-15pt}
        \caption{\textit{vision-block10}}
    \end{subfigure}
    \hfill
    \begin{subfigure}[t]{0.3\textwidth}
        \centering
        \includegraphics[width=\linewidth]{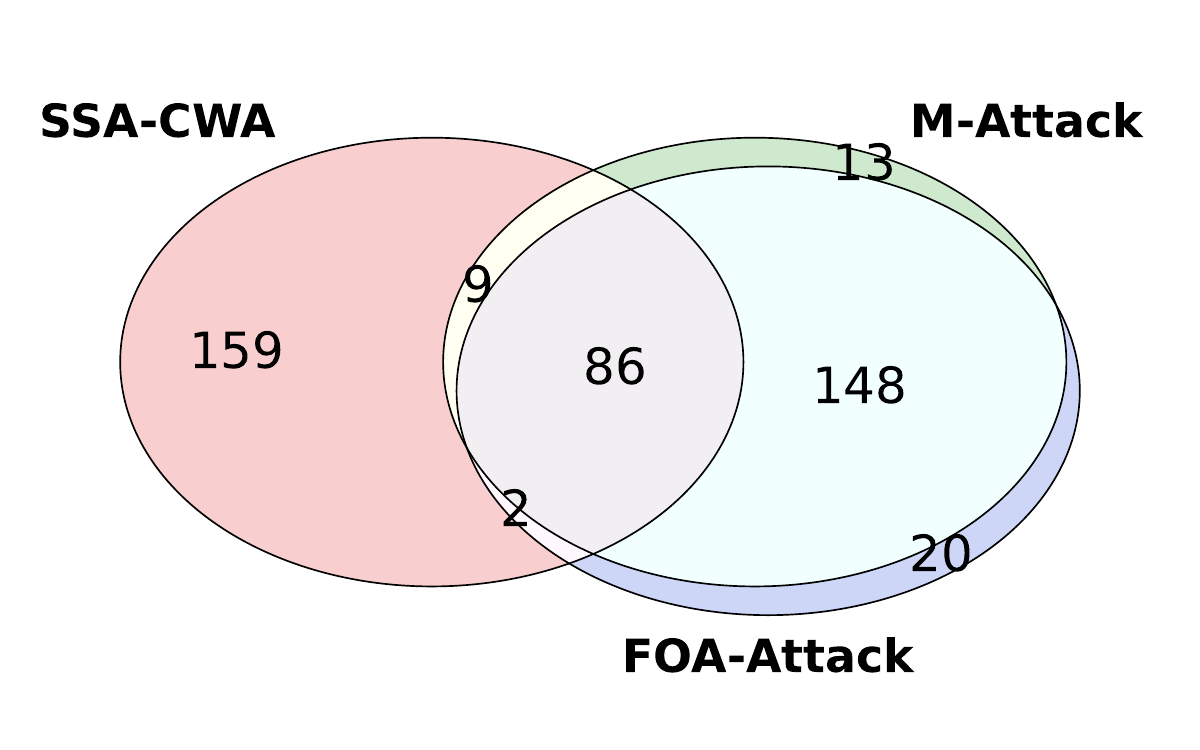}
        \vspace{-15pt}
        \caption{\textit{projection-mlp2}}
    \end{subfigure}
    \caption{Shared feature overlap across attack methods, illustrated by Venn diagrams of the top-256 features extracted from the NIPS17 dataset at three different layer locations.}
    \vspace{-5pt}
    \label{fig:feature-venn-cross-attack}
\end{figure}

Figure~\ref{fig:feature-venn-cross-attack} visualizes the overlap of activated features across different attack methods.
We observe that at the \textit{vision-block0} layer, a large number of features are shared across different attacks, with M-Attack and FOA-Attack exhibiting nearly complete overlap, likely due to their similar design and strong resemblance in generated perturbations.
As the layer depth increases, the overlap between SSA-CWA and the other two attacks gradually decreases, whereas M-Attack and FOA-Attack continue to exhibit a high degree of feature overlap.
This suggests that some features correspond to more local perturbation patterns, while others reflect more global feature shifts. Such a trend is consistent with our hypothesis in Sec.~\ref{main-results}.

\begin{table}[t]
\centering
\caption{Cross-attack results.}
\label{tab:cross-attack}
\small
\resizebox{0.89\linewidth}{!}{%
\begin{tabular}{llrrrrrrrrr}
\toprule
\multicolumn{2}{c}{} & \multicolumn{3}{c}{NIPS17} & \multicolumn{3}{c}{LLaVA} & \multicolumn{3}{c}{Medical} \\
\cmidrule(lr){1-2} \cmidrule(lr){3-5} \cmidrule(lr){6-8} \cmidrule(lr){9-11}
Transfer Setting & Method & \multicolumn{1}{c}{P} & \multicolumn{1}{c}{R} & \multicolumn{1}{c}{F1} & \multicolumn{1}{c}{P} & \multicolumn{1}{c}{R} & \multicolumn{1}{c}{F1} & \multicolumn{1}{c}{P} & \multicolumn{1}{c}{R} & \multicolumn{1}{c}{F1} \\
\midrule
\multirow{6}{*}{SSA-CWA$\rightarrow$M-Attack} & Dense & \underline{100.0} & 4 & 7.6 & 89.7 & 35 & 50.3 & 98.8 & 88 & 93.1 \\
 & Dense (Ensemble) & 99.0 & \underline{100} & \textbf{99.5} & \underline{99.0} & \underline{100} & \textbf{99.5} & \underline{100.0} & 88 & 93.6 \\
 & PIP & 97.2 & 71 & 82.0 & 97.8 & 93 & 95.3 & 97.8 & \underline{90} & \textbf{93.7} \\
 & SAEgis (vision-block0) & \underline{100.0} & 91 & 95.2 & 98.0 & 98 & 98.0 & 97.6 & 83 & 89.7 \\
 & SAEgis (projection-mlp2) & 92.1 & 35 & 50.7 & 96.6 & 58 & 72.5 & 96.7 & 30 & 45.8 \\
 & SAEgis (Ensemble) & \underline{100.0} & 99 & \textbf{99.5} & 98.0 & 99 & 98.5 & 97.8 & \underline{90} & \textbf{93.7} \\
\midrule
\multirow{6}{*}{SSA-CWA$\rightarrow$FOA-Attack} & Dense & \underline{100.0} & 3 & 5.8 & 90.7 & 39 & 54.5 & 98.8 & 89 & 93.6 \\
 & Dense (Ensemble) & 99.0 & \underline{100} & \textbf{99.5} & \underline{99.0} & \underline{100} & \textbf{99.5} & \underline{100.0} & \underline{93} & \textbf{96.3} \\
 & PIP & 96.9 & 64 & 77.1 & 97.8 & 89 & 93.1 & 97.8 & 92 & 94.8 \\
 & SAEgis (vision-block0) & \underline{100.0} & 90 & 94.7 & 98.0 & 98 & 98.0 & 97.6 & 83 & 89.7 \\
 & SAEgis (projection-mlp2) & 92.5 & 37 & 52.8 & 96.5 & 56 & 70.8 & 96.5 & 28 & 43.4 \\
 & SAEgis (Ensemble) & \underline{100.0} & 98 & 98.9 & 98.0 & 98 & 98.0 & 97.8 & 91 & 94.3 \\
\bottomrule
\end{tabular}}
\end{table}
\begin{table}[t]
\centering
\vspace{-10pt}
\caption{Overall results.}
\label{tab:overall}
\small
\resizebox{0.72\linewidth}{!}{%
\begin{tabular}{lrrrrrrrrr}
\toprule
\multicolumn{1}{c}{} & \multicolumn{3}{c}{In-domain} & \multicolumn{3}{c}{Cross-domain} & \multicolumn{3}{c}{Cross-attack} \\
\cmidrule(lr){1-1} \cmidrule(lr){2-4} \cmidrule(lr){5-7} \cmidrule(lr){8-10} Method & \multicolumn{1}{c}{P} & \multicolumn{1}{c}{R} & \multicolumn{1}{c}{F1} & \multicolumn{1}{c}{P} & \multicolumn{1}{c}{R} & \multicolumn{1}{c}{F1} & \multicolumn{1}{c}{$\Delta$ P} & \multicolumn{1}{c}{$\Delta$ R} & \multicolumn{1}{c}{$\Delta$ F1} \\
\midrule
Dense & \underline{98.3} & 88.4 & 93.0 & 67.8 & 92.5 & 75.1 & -1.7& -43.3& -40.9\\
 Dense (Ensemble) & 97.6 & \underline{98.6} & \textbf{98.0} & 71.4 & \underline{99.9} & 82.2 & \underline{+2.5}& -1.5& \textbf{+0.4}\\
 PIP & 97.8 & 92.4 & 95.0 & \underline{96.8} & 70.4 & 79.4 & -0.2 &-7.5& -4.7\\
 SAEgis & 98.1 & 89.9 & 93.7 & 93.3 & 85.3 & 88.4 & -3.0& -46.6& -36.3\\
 SAEgis (Ensemble) & 97.7 & 97.0 & 97.3 & 93.1 & 96.2 & \textbf{94.4} & +1.3& \underline{-1.0}& +0.1 \\
\bottomrule
\end{tabular}}
\end{table}

\begin{figure}[H]
    \centering
    \begin{subfigure}[t]{0.32\textwidth}
        \centering
        \includegraphics[width=\linewidth]{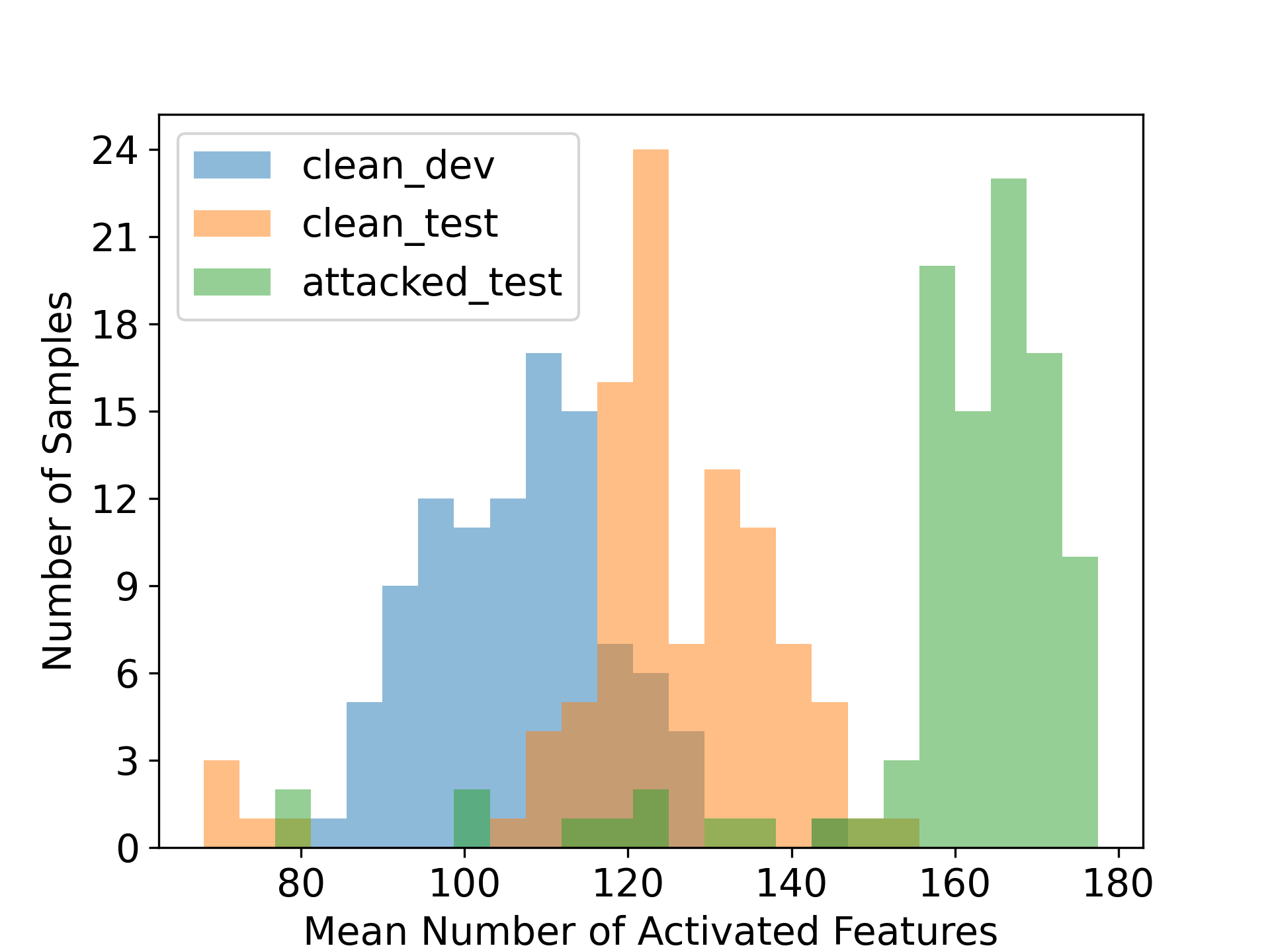}
        \caption{Case of \textbf{low precision} (P = 69.1\%), SSA-CWA is applied under cross-domain (LLaVA $\rightarrow$ Medical).}
    \end{subfigure}
    \hfill
    \begin{subfigure}[t]{0.32\textwidth}
        \centering
        \includegraphics[width=\linewidth]{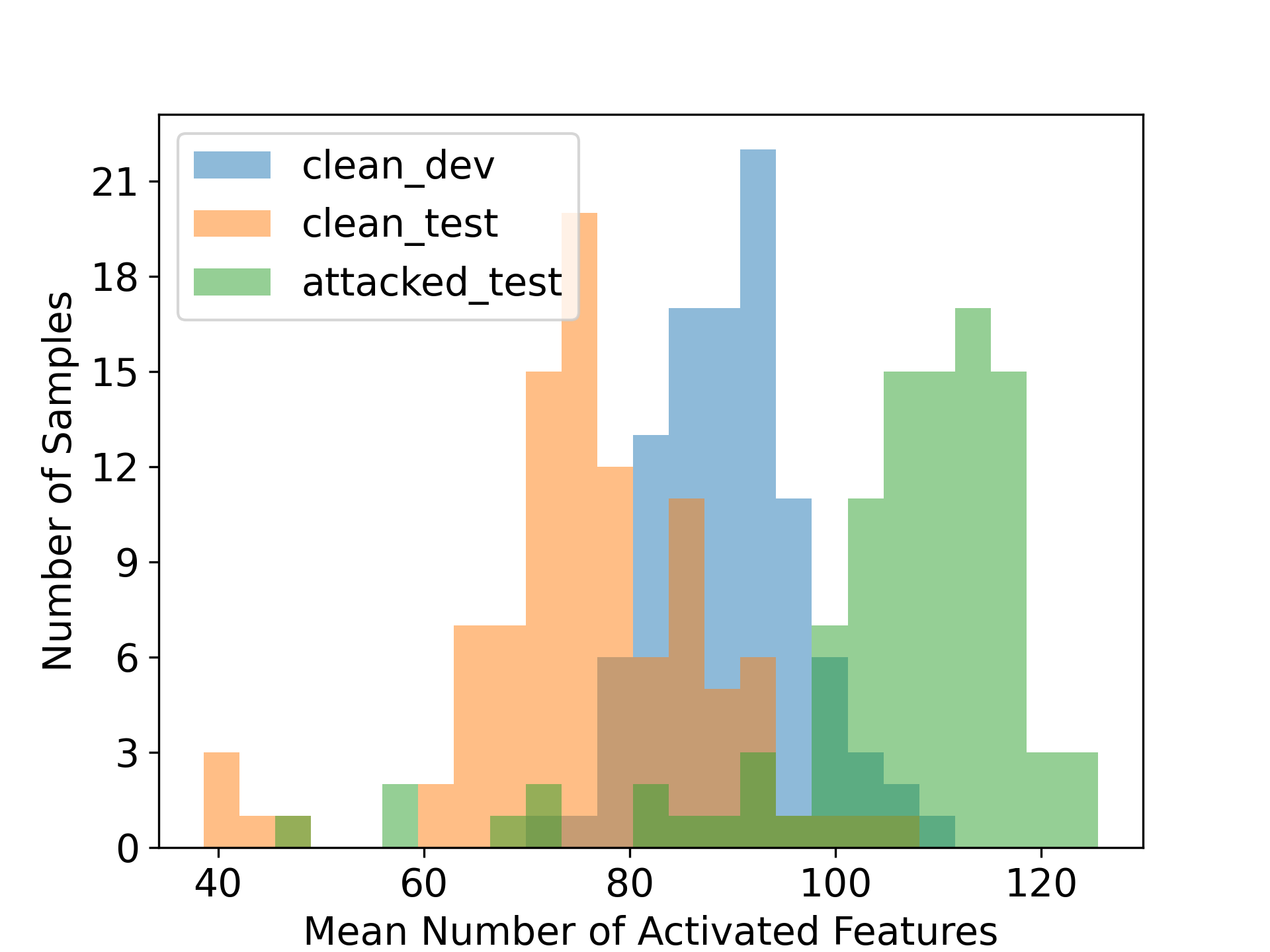}
        \caption{Case of \textbf{low recall} (R = 66\%), M-Attack is applied under cross-domain (LLaVA $\rightarrow$ Medical).}
    \end{subfigure}
    \hfill
    \begin{subfigure}[t]{0.32\textwidth}
        \centering
        \includegraphics[width=\linewidth]{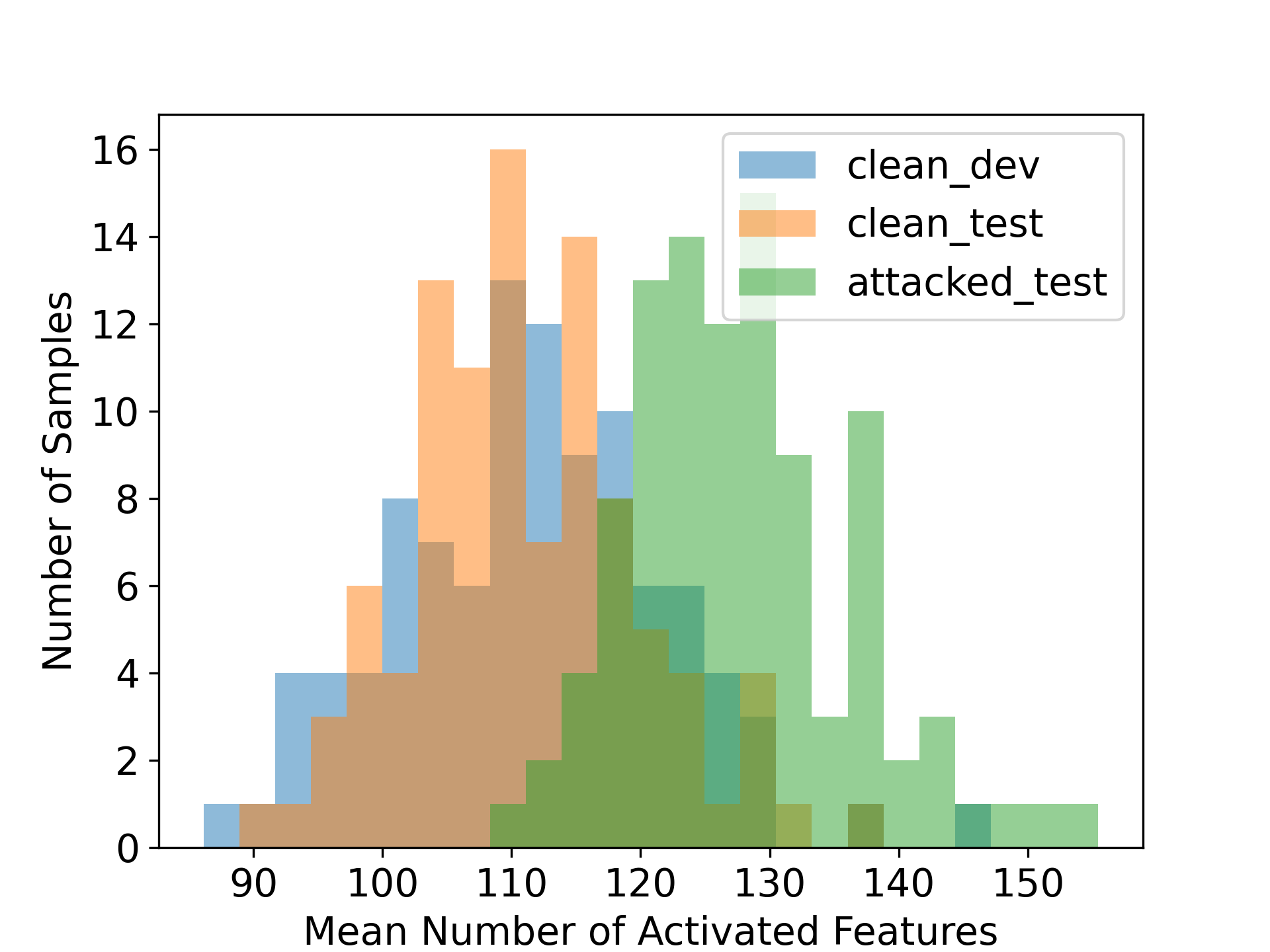}
        \caption{Case of a \textbf{mixed distribution} (R = 35\%), NIPS17 is used under cross-attack (SSA-CWA $\rightarrow$ M-Attack).}
    \end{subfigure}
    \caption{Distributions of the number of activated features for clean and adversarial images, averaged over image tokens, with all results computed at the \textit{projection-mlp2} layer.}
    \label{fig:feature-distribution}
    \vspace{-5pt}
\end{figure}

\paragraph{Failure cases from feature activation distributions}
Ideally, SAEgis should activate substantially more features for adversarial images than for clean ones, resulting in well-separated distributions of activation counts.
However, under cross-domain or cross-attack shifts, the test data distribution may diverge from that of the training and development sets used for threshold calibration, resulting in degraded performance.
In Figure~\ref{fig:feature-distribution}, we present three representative failure cases with low accuracy.
In (a), the distribution of clean test images shifts to the right, activating more attack-relevant features, which makes the threshold estimated from clean dev data too low and causes some clean images to be misclassified, reducing precision.
In (b), the opposite occurs: the test distribution shifts to the left, leading to an overly high threshold that misclassifies some adversarial images as clean, thus lowering recall.
In (c), although the distributions of clean dev and clean test data largely overlap, cross-attack shifts induce different global feature activations, causing the distributions of clean and adversarial images to be inherently mixed, making it difficult to separate them regardless of the threshold choice.
\subsection{Ablation Study}
\vspace{-5pt}
\label{analysis:ablation}

We conduct additional ablation studies to evaluate the impact of multi-layer ensembling and the number of adversarial samples used for feature selection.
As shown in Table~\ref{tab:ensemble-medical}, although \textit{vision-block0} achieves strong performance in the cross-attack setting and deeper layers exhibit progressively lower recall, incorporating \textit{projection-mlp2} into the ensemble still yields additional gains, with the three-layer ensemble performing the best overall.
This suggests that, similar to adversarial attacks, effective detection benefits from jointly modeling both local and global features.
Furthermore, Figure~\ref{fig:few-shot} shows that using as few as 10 adversarial samples (together with 800 clean images) already achieves reasonably strong performance (around 80\% F1), highlighting the practicality of SAEgis in black-box settings where access to adversarial data is limited.

\begin{minipage}[t]{0.48\linewidth}
    \vspace{0pt}
    \centering
    \captionof{table}{Ensemble results under cross-attack settings on the Medical dataset.}
    \label{tab:ensemble-medical}
    \small
    \resizebox{\linewidth}{!}{%
    \begin{tabular}{llrrr}
    \toprule
    \multicolumn{2}{c}{} & \multicolumn{3}{c}{Medical} \\
    \cmidrule(lr){1-2} \cmidrule(lr){3-5}
    Transfer Setting & Layer & \multicolumn{1}{c}{P} & \multicolumn{1}{c}{R} & \multicolumn{1}{c}{F1} \\
    \midrule
    \multirow{7}{*}{\makecell{SSA-CWA \\ $\rightarrow$ M-Attack}} & vision-block0 & 97.6 & 83 & 89.7 \\
     & vision-block10 & \underline{100.0} & 56 & 71.7 \\
     & projection-mlp2 & 96.7 & 30 & 45.8 \\
     & Ensemble (vis0 + vis10) & 98.7 & 79 & 87.7 \\
     & Ensemble (vis0 + proj) & 96.7 & \underline{90} & 93.2 \\
     & Ensemble (vis10 + proj) & 97.7 & 88 & 92.6 \\
     & Ensemble (vis0 + vis10 + proj) & 97.8 & \underline{90} & \textbf{93.7} \\
    \bottomrule
    \end{tabular}}
\end{minipage}
\hfill
\begin{minipage}[t]{0.48\linewidth}
    \vspace{0pt}
    \centering
    \includegraphics[width=\linewidth]{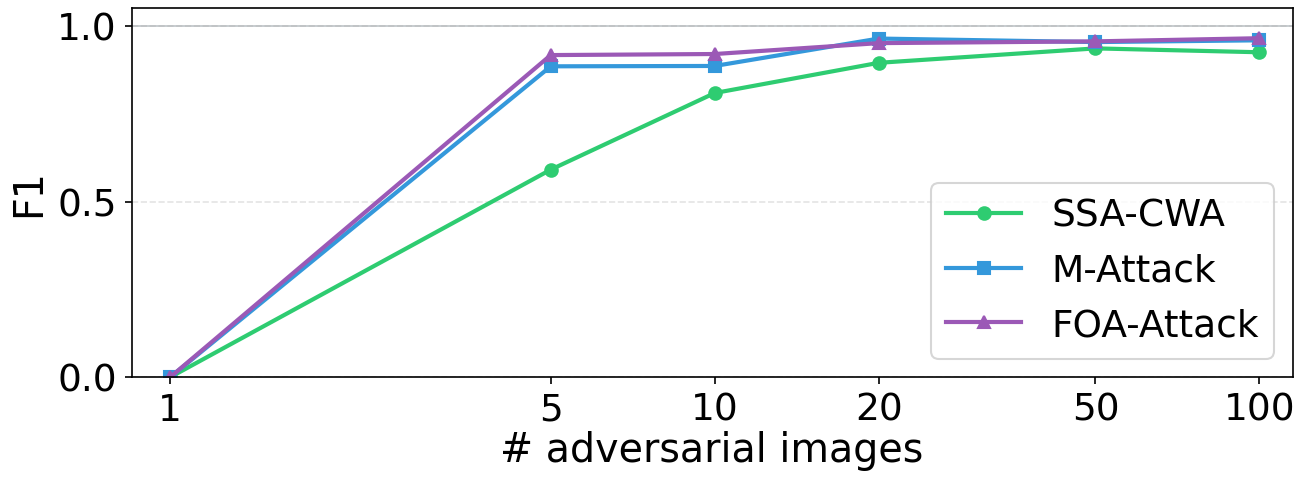}
    \vspace{-15pt}
    \captionof{figure}{F1 vs. number of adversarial samples for feature selection, under cross-domain (Medical $\rightarrow$ NIPS17) at \textit{projection-mlp2}.}
    \label{fig:few-shot}
\end{minipage}
\section{Conclusion}
In this work, we introduced SAEgis, a simple and effective framework for adversarial attack detection in vision-language models by leveraging sparse autoencoders as plug-and-play modules.
Without requiring adversarial training, SAEgis identifies attack-relevant features and achieves strong performance across in-domain, cross-domain, and cross-attack settings.
We further showed that fusing signals across layers improves stability and performance, with each layer contributing complementary representations at different levels of abstraction.
Overall, our results suggest that sparse latent features provide a practical and reliable foundation for enhancing the safety of real-world VLM systems.
\newpage
\section{Acknowledgements}
We thank Koshiro Aoki, Sebastian Zwirner, and Wentao Hu for their helpful discussions.
\bibliographystyle{abbrvnat_custom}
\bibliography{ref}

\end{document}